\documentclass[preprint,12pt]{elsarticle}

\usepackage{amssymb}
\usepackage{amsmath}
\usepackage{bbm}
\usepackage{multirow}
\usepackage{booktabs}
\usepackage{xcolor}
\usepackage{hyperref}
\usepackage[ruled,vlined]{algorithm2e}
\usepackage{amsmath}
\usepackage{tablefootnote}
\usepackage{makecell}
\usepackage{array}
\newcolumntype{C}[1]{>{\centering\arraybackslash}m{#1}}

\journal{Engineering Applications of Artificial Intelligence}

\begin{document}

\begin{frontmatter}



\title{One-shot federated unsupervised domain adaptation with Smoothed Knowledge Distillation and teacher refinement}



\author[label1]{Ali Abedi}
\author[label1]{Q. M. Jonathan Wu}
\author[label1]{Ning Zhang}

\affiliation[label1]{organization={University of Windsor, Department of Electrical and Computer Engineering}}

\author[label2]{Farhad Pourpanah} 

\affiliation[label2]{organization={Queen's University, Department of Electrical and Computer Engineering}}

\begin{abstract}
Federated unsupervised domain adaptation (FUDA) addresses the challenge of adapting models to an unlabeled target domain using decentralized source domains while preserving data privacy. However, existing FUDA methods typically require multiple communication rounds, rely on complex aggregation strategies, and often struggle with noisy pseudo labels and inconsistent source knowledge.
To address these challenges, we propose SKD-ETR, a novel one-shot FUDA framework that combines Smoothed Knowledge Distillation (SKD) and Exponential moving average-based Teacher Refinement (ETR).
SKD trains a student model on the target domain using smoothed soft pseudo labels generated by the ensemble of source models. This reduces overconfidence, mitigates noise, and improves robustness. ETR further refines each source model by interpolating its parameters toward the student via exponential moving average (EMA), thereby transferring target-domain knowledge back to the teachers. This bidirectional refinement enhances pseudo-label quality and student generalization without additional communication overhead. SKD-ETR eliminates complex aggregation by initializing the student model randomly and performing a single-round distillation process. Extensive experiments on OfficeHome, Office-Caltech, and DomainNet demonstrate that SKD-ETR achieves competitive performance while being communication- and computation-efficient, and robust under noisy supervision. The code for this paper is available at \url{https://github.com/A-Abedi/SKD_ETR}.
\end{abstract}


\end{frontmatter}



\section{Introduction} 
Federated learning (FL) is a decentralized learning framework where multiple clients collaboratively train a shared global model without exchanging their raw data \cite{wen_survey_2023, shenaj_federated_2023, zhang_survey_2021}. This approach is critical in scenarios where data is distributed across multiple centers and cannot be shared due to privacy regulations, such as in medical imaging \cite{wen_survey_2023, guan_federated_2024, wang_bm-pda_2025}. However, FL faces two major challenges in real-world deployments. First, conventional FL methods rely on multiple rounds of communication between the server and clients to form the global model \cite{mcmahan_communication-efficient_2017}, which results in high communication overhead and an increased risk of privacy leakage during transmission \cite{guha_one-shot_2019, zhang_privacyeafl_2023, zhou_anonymous_2023}. Second, in practice, the data distributions across clients are often non-identical and domain-specific. This distributional heterogeneity, also referred to as domain shift \cite{sun_survey_2015, li_comprehensive_2024, wang_deep_2018, tistarelli_domain_2023}, can severely degrade the generalization performance of the global model when applied to unseen target domains.

To overcome these limitations, recent studies have proposed strategies to improve both scalability and generalization in FL. One-shot FL \cite{guha_one-shot_2019, su_one-shot_2023, beitollahi_parametric_2024} offers a communication-efficient alternative by performing only a single communication round. This design reduces bandwidth requirement and minimizes the risk of privacy leakage, making it suitable for resource-constrained and privacy-sensitive settings \cite{chen_fedbip_2025}. Unlike traditional multi-round FL, which frequently exchanges parameters between server and clients, one-shot FL eliminates repeated exposure of intermediate model states and thus reduces the attack surface.

Federated unsupervised domain adaptation (FUDA) \cite{peng_federated_2019} extends FL to address domain shift by adapting a model trained on distributed labeled source domains to an unlabeled target domain while maintaining data privacy. Each client trains locally on its labeled source domain, and the server aggregates knowledge from source models to generalize to an unlabeled target domain \cite{hong_federated_2021, xiao_unsupervised_2024, pourpanah_federated_2025, abedi_one-shot_2025}. FUDA can be viewed as a decentralized and privacy-preserving extension of multi-source-free domain adaptation (MSFDA) \cite{fang_source-free_2024}, which addresses the challenge of domain adaptation without access to source data by leveraging pretrained source models. MSFDA itself builds upon the multi-source unsupervised domain adaptation (MSUDA) \cite{sun_survey_2015} framework, which assumes access to labeled data from multiple source domains during the adaptation on an unlabeled target domain. While MSUDA relies on centralized access to both source and target data, and MSFDA assumes access to all source models in a central location, such assumptions are often impractical due to privacy and communication constraints. This motivates the development of FUDA. 

Existing FUDA methods \cite{yi_fdac_2025, peng_federated_2019, hong_federated_2021, xiao_unsupervised_2024, abedi_one-shot_2025} have achieved notable progress but still face several fundamental limitations. Source-to-target knowledge transfer remains unstable because consensus voting or confidence thresholds may either discard valuable signals or propagate noisy supervision. Aggregation strategies are often heuristic, which can introduce bias. Moreover, many approaches are communication-inefficient, limiting their scalability in real federated settings. As summarized in Table~\ref{table:comp_intro}, these issues highlight the gap between existing multi-round and one-shot designs and motivate the need for a more communication-efficient and robust framework that can stabilize pseudo-labels while reducing reliance on costly aggregation mechanisms \cite{tang_learning_2022, tang_m3net_2023}.

These limitations motivate our one-shot FUDA framework, SKD-ETR, which integrates Smoothed Knowledge Distillation (SKD) and Exponential moving average-based Teacher Refinement (ETR).
The framework builds on Knowledge Distillation (KD) and consists of two key modules. The first module (SKD) trains a randomly initialized student model from scratch on the unlabeled target domain using soft pseudo labels generated by an ensemble of independently trained source models. To handle the impact of noise and uncertainty inherent in these teacher-generated pseudo labels, we incorporate label smoothing \cite{soltany_federated_2025} into the Kullback-Leibler (KL) divergence loss. This acts as an implicit regularizer and improves the robustness and generalization of the student model \cite{tang_blockmix_2020}. The second module (ETR) updates each source model by gradually interpolating its parameters toward the student through an exponential moving average (EMA) mechanism. This process allows the teachers to absorb target-specific knowledge and produce more robust and target-aware pseudo labels, thereby enhancing adaptation performance. The effectiveness of SKD-ETR is validated on three standard benchmarks: OfficeHome \cite{venkateswara_deep_2017}, Office-Caltech \cite{gong_geodesic_2012}, and DomainNet \cite{peng_moment_2019}.

\begin{table}[tb]
\caption{Comparison of representative one-shot and multi-round FUDA methods. $R$ denotes the number of rounds and $M$ the number of clients.}
\label{table:comp_intro}
\centering
\resizebox{\textwidth}{!}{
\begin{tabular}{l|cC{4cm}cC{4cm}C{4cm}}
\toprule
Model & FL rounds & Communication & $\sim$ Communication cost & Aggregation & Adaptation \\ \midrule
FDAC \cite{yi_fdac_2025} & R & Local models & $R \times M\times$ 85M Params & FedAvg & \makecell{Domain alignment +\\semantic matching}  \\
FADA \cite{peng_federated_2019} & R & Frequent gradient sharing & $R \times M\times$ 45M Params & Source gradients aggregation & Adversarial feature disentanglement \\
FADE \cite{hong_federated_2021} & R & \makecell{Local models +\\Global Model} & $R \times M\times$ 45M Params & FedAvg & Adversarial debiasing \\
KD3A \cite{feng_kd3a_2021} & R & \makecell{Local models +\\Global Model} & $R \times M\times$ 23M Params & Consensus focus reweighting & \makecell{Confidence gating +\\knowledge-vote ensemble}  \\
FDKD \cite{xiao_unsupervised_2024} & R &  \makecell{Local models +\\Global Model} & $R \times M\times$ 23M Params & No aggregation step & \makecell{Confidence gating +\\knowledge-vote ensemble}  \\
SEA+MSPL \cite{abedi_one-shot_2025} & 1 & \makecell{Bottleneck and head \\of local models} & $1 \times M\times$ 4M Params & Scaled-entropy weighting & Multi-source pseudo labeling + SSCE  \\
SKD-ETR & 1 & \makecell{Bottleneck and head \\of local models} & $1 \times M\times$ 4M Params & No aggregation step & Smoothed knowledge distillation \\
\bottomrule
\end{tabular}
}
\end{table}

Our main contributions in this research are summarized as follows:
\begin{itemize}
    \item We propose a one-shot FUDA framework that leverages KD to adapt a student model using only unlabeled target data and pretrained source models.
    \item We introduce SKD to regularize the learning process by mitigating noisy pseudo labels through label smoothing within the KL divergence loss, which makes the adaptation process more robust.
    \item We develop ETR, supported by theoretical justification, to update source models using the student which, enables teachers to incorporate target-domain knowledge and generate more reliable pseudo-labels without requiring additional communication.
    \item  We conduct comprehensive evaluations on multiple standard benchmarks and demonstrate that our method consistently outperforms existing approaches, confirming its robustness and effectiveness in diverse domain adaptation scenarios.
\end{itemize}

The rest of this paper is organized as follows. Section~\ref{sec:related_works} provides a review of MSUDA, FUDA, and KD. Section~\ref{sec:methodology} presents details of our proposed SKD-ETR model. Details of the experimental setting are presented in Section~\ref{sec:implementation_details}. Section~\ref{sec:results} provides the experimental results, ablation studies, and sensitivity analyses. Finally, Section~\ref{sec:conclusion} concludes the paper and outlines directions for future work.

\section{Related Work}
\label{sec:related_works}
In this section, we review related works across several areas relevant to our approach, including MSUDA, FUDA, and KD.

\subsection{Multi-source Unsupervised Domain Adaptation}
SSD \cite{li_multidomain_2024} is a MSUDA method that selects informative source samples and ranks source domains through a two-step distillation process using pseudo-labeled target data. A domain discriminator assigns weights and identifies the dominant source domain for final prediction. SImpAl \cite{venkat_your_2020} enforces classifier agreement on pseudo-labeled target samples, which enables implicit feature alignment without additional alignment losses. MDAMA \cite{dayal_leveraging_2025} aligns the target domain with a mixture of source domains using a classifier-aware margin-based discrepancy. These methods assumes access to all source data during adaptation, which is often impractical due to privacy or communication limits.

MSFDA \cite{fang_source-free_2024} overcomes this limitation by adapting using only pretrained source models, without source data access. SHOT \cite{liang_we_2020} freezes the source classifier and optimizes a target-specific feature encoder using information maximization and self-supervised pseudo-labeling. DECISION \cite{ahmed_unsupervised_2021} performs multi-source-free domain adaptation by learning optimal source model weights and adapting only feature extractors using information maximization and pseudo-labeling strategies. TransDA \cite{yang_transformer-based_2021} injects a transformer module into the feature extractor and applies self-knowledge distillation with pseudo-labels to guide attention toward object regions in the target domain.

\subsection{Federated Unsupervised Domain Adaptation}

Unlike MSFDA, which assumes centralized access to all source models, FUDA distributes them across clients and supports privacy-preserving adaptation with unlabeled target data.
FADE \cite{hong_federated_2021} uses adversarial debiasing to align feature distributions across non-iid clients without requiring group labels or data sharing. FDKD \cite{xiao_unsupervised_2024} uses confidence-weighted KD with adaptive teacher weighting and feature-level alignment to improve generalization on the target domain. KD3A \cite{feng_kd3a_2021} distills knowledge from source model predictions using consensus voting and reweights domains based on their contribution to reduce negative transfer. FDAC \cite{yi_fdac_2025} applies domain- and category-level contrastive learning to align features and preserve class semantics across source domains. SEA+MSPL \cite{abedi_one-shot_2025} introduces one-shot FUDA by aggregating source models using scaled entropy-based aggregation and adapting the global model via smoothed pseudo labels from source predictions.

\subsection{Knowledge Distillation}
Knowledge distillation (KD) \cite{hinton_distilling_2015, moslemi_survey_2024, tang_connecting_2025}, which was originally introduced as a model compression technique to transfer knowledge from a large and well-performing teacher model to a smaller student model, has shown promising results in domain adaptation tasks. KD3A \cite{feng_kd3a_2021} uses a knowledge voting scheme that aggregates confident predictions from multiple teachers to form a consensus signal for the student. FDKD \cite{xiao_unsupervised_2024} introduces confidence-based multi-teacher distillation by adaptively weighting teachers based on their agreement with smoothed ensemble predictions. SSD \cite{li_multidomain_2024} performs dual distillation by weighting both source models and individual samples based on cross-domain consistency and reliability during target training. DECISION \cite{ahmed_unsupervised_2021} trains a student by combining weighted information maximization with refined pseudo labels from fixed classifiers. TransDA \cite{yang_transformer-based_2021} applies self-distillation using soft pseudo labels to improve attention alignment in target adaptation. 

\section{Methodology}
\label{sec:methodology}

\subsection{Problem Formulation}
Assume there are $M$ clients, each corresponding to a unique source domain $\mathcal{D}_{i}^{S} = \{(x_{ij}, y_{ij})\}_{j=1}^{N_i}$, where $x_{ij} \in \mathcal{X}_i$ and $y_{ij} \in \mathcal{Y}_i$ denote the $j$-th data sample and its associated label for the $i$-th client. These source domains exhibit distributional heterogeneity, i.e., $\mathcal{D}_i^S \sim P_i(X, Y)$ and $P_i \neq P_k$ for $i \ne k$, reflecting differences in data distributions across clients. Each client trains a local model $f_i$ on its private dataset by minimizing the standard cross-entropy loss, as follows:
\begin{equation} \min \mathcal{L}_i(f_i) = \frac{1}{N_i} \sum_{j=1}^{N_i} \ell(f_i(x_{ij}), y_{ij}), \end{equation}
where $\ell(\cdot, \cdot)$ is the cross-entropy loss function. Importantly, due to privacy constraints, raw data, features, or any statistics are not shared between clients or with a central server.

The objective is to leverage the knowledge encoded in these independently trained source models to learn a student model $f_s$ that performs well on an unlabeled target domain ${\mathcal{D}}^T = {x_j^T}_{j=1}^{N_T}$, with $x_j^T \in \mathcal{X}^T$. The student model is trained from scratch using supervision derived from the source models. To adapt $f_s$ to the target domain, we minimize the following loss:
\begin{equation} \min \mathcal{L}_{\text{target}}(f_s) = \frac{1}{N_T} \sum_{j=1}^{N_T} \ell(f_s(x_j^T), \bar{\mathcal{D}}^T), \end{equation}
where $\bar{\mathcal{D}}^T$ represents the knowledge transferred from the source models, i.e., soft pseudo-labels in this study.

\subsection{Model Overview}

\begin{figure*}[t]
\centering
\begin{minipage}{\linewidth}
\includegraphics[width=\textwidth]{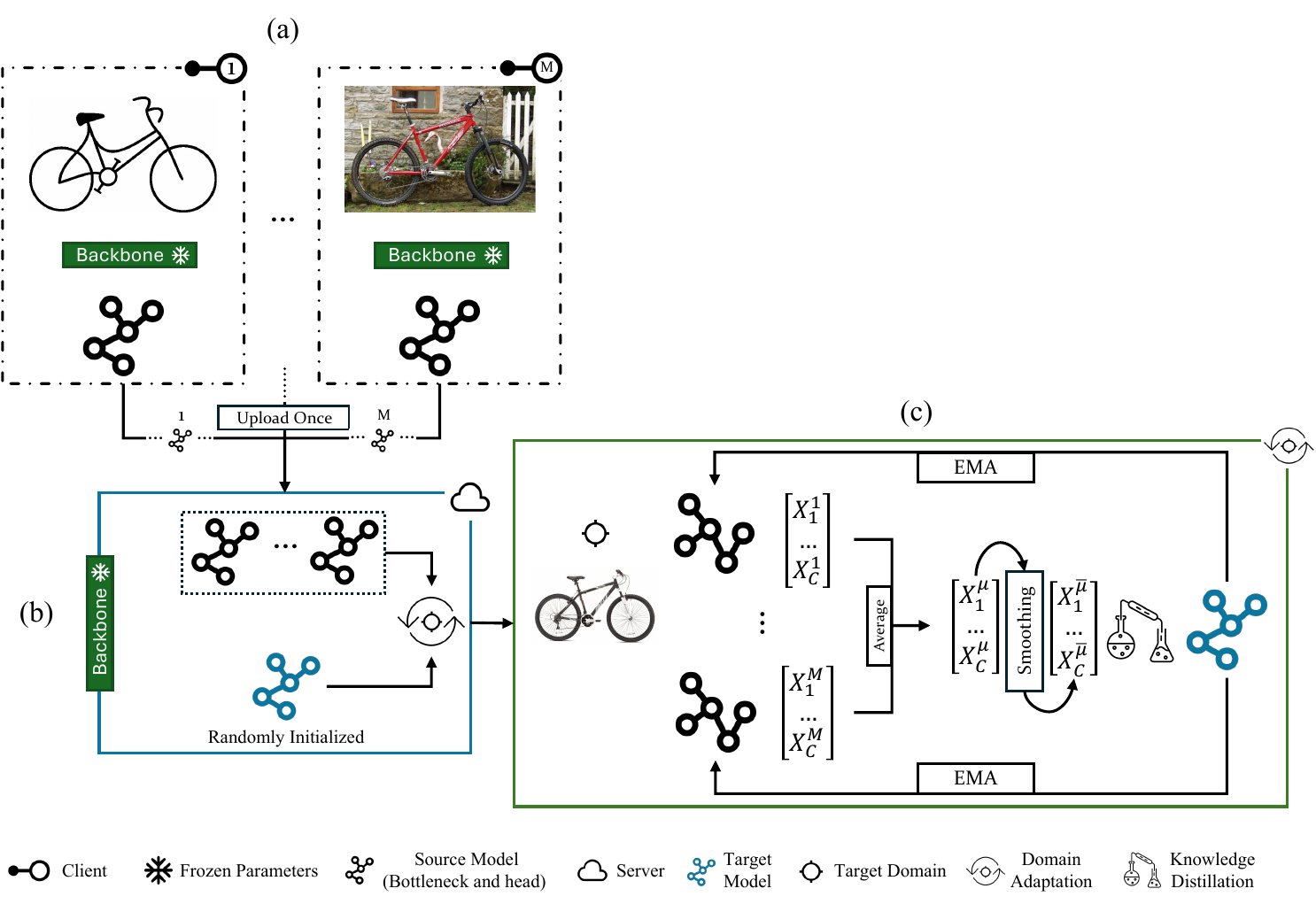}
\caption{Overview of SKD-ETR.
\textbf{(a)} Each client independently trains a source model on its private labeled dataset drawn from a distinct distribution.
\textbf{(b)} The server initializes a randomly initialized student model for adaptation.
\textbf{(c)} The source models act as teachers to generate smoothed pseudo labels for the unlabeled target-domain data, which are used to train a student model via KD with KL divergence. Then each source model is refined using an exponential moving average of the student model’s parameters to improve generalization without additional communication.\protect\footnotemark[1]}
\label{fig:model_overview}
\end{minipage}
\end{figure*}

\footnotetext[1]{Distillation icons created by Prosymbols Premium - Flaticon}

Figure~\ref{fig:model_overview} presents an overview of our proposed SKD-ETR method for the one-shot FUDA. Each client independently trains a source model on its private labeled dataset drawn from a distinct distribution. To preserve privacy, only trainable parameters of each model are shared with the server. The server then uses a randomly initialized student model for adaptation. The domain adaptation is proceed in two stages: (1) SKD, where the source models act as teachers to guide the training of a student model on the unlabeled target domain via KD (as described in Section~\ref{sec:skd}); and (2) ETR, where each teacher model is updated via an EMA of the student’s parameters. This refinement improves supervision quality to generate more robust pseudo-labels without additional communication rounds. Since the goal is to obtain a single student model that performs well on the unlabeled target domain, the refined teachers are not used after training (explained in Section~\ref{sec:etr}).

\subsection{Model Architecture}
\label{sec:architecture}
Our model follows a modular architecture influenced by recent works in lightweight domain adaptation~\cite{abedi_one-shot_2025}, and consists of three components: a backbone, a bottleneck, and a head.
The backbone is a frozen pre-trained DINOv2~\cite{oquab_dinov2_2024}, based on ViT~\cite{dosovitskiy_image_2021}, used for extracting robust features from unlabeled data. The bottleneck consists of multiple fully connected layers that transform features from the frozen backbone into a more transferable representation~\cite{abedi_euda_2024}, while the head is a single fully connected layer that maps these transformed features to class logits. The model is defined as:
\begin{equation}
  f_i(x_i) = h_{\phi_i}\big(g_{\psi_i}(b_{\omega}(x_i))\big),
  \label{eq:overal_arch}
\end{equation}
where $b_{\omega}$ is the frozen backbone with parameters $\omega$, and $g_{\psi_i}$ and $h_{\phi_i}$ are the trainable bottleneck and head, respectively. After training, clients send only the bottleneck and head parameters $\theta_i = \{\psi_i, \phi_i\}$ to the server for aggregation.

\subsection{Source Model Training and Server-Side Aggregation}
\label{sec:smt_ssa}
Each client independently trains a source model $f_i$ on its private labeled dataset $\mathcal{D}_i^S$, where $x_{ij}$ is an input sample and $y_{ij} \in \{1, \dots, C\}$ is its corresponding label. No data, features, or gradients are shared across clients. The local training objective minimizes the standard cross-entropy loss:
\begin{equation}
    \label{eq:cross_ent}
    \mathcal{L}_i(f_i) = - \frac{1}{N_i} \sum_{j=1}^{N_i} \sum_{c=1}^{C} \mathbbm{1}[y_{ij} = c] \log p_i(x_{ij})_c,
\end{equation}
where $p_i(x_{ij}) = \text{softmax}(f_i(x_{ij}))$ denotes the predicted class probabilities, and $\mathbbm{1}[y_{ij} = c]$ is the indicator function for the ground-truth label.

Once all $M$ clients complete local training, their parameters $\{\theta_i\}_{i=1}^{M}$ are transmitted to the server. Instead of aggregating the source models, the server begins the domain adaptation process by initializing the student model $f_s$ with random parameters $\theta_s$. This design avoids complex model aggregation and prevents bias introduced by domain-specific source representations. Each source model converges to a solution shaped by its own domain-specific distribution, which would bias the student’s initialization if used directly and steer its optimization toward source-dependent minima. A randomly initialized student provides a neutral starting point that is not influenced by any individual source. The subsequent training on the unlabeled target domain via KD allows the student’s representation to be guided primarily by target-informed pseudo labels produced by the teacher ensemble. This design supports the goal of learning a more domain-invariant initialization and reduces the influence of source-domain bias on the student’s parameter trajectory.

\subsection{Smoothed Knowledge Distillation}
\label{sec:skd}
We propose SKD to adapt a student model $f_s$ to the target domain $\mathcal{D}^T$ by leveraging predictions from multiple independently trained source models $\{f_i\}_{i=1}^{M}$. Instead of relying on hard pseudo labels, SKD uses temperature-scaled soft targets to improve robustness against noisy predictions and domain discrepancies.
For each unlabeled sample $x_j^T$ in the target domain, each source model $f_i$ generates a logit vector, which is then converted into a probability distribution via softmax:
\begin{equation}
    \label{eq:client_softmax}
    p_i(x_j^T) = \text{softmax}(f_i(x_j^T)) \in \mathbb{R}^C.
\end{equation}

These individual distributions are averaged to obtain an ensemble prediction:
\begin{equation}
    \label{eq:ensemble_pred}
    \bar{p}(x_j^T) = \frac{1}{M} \sum_{i=1}^{M} p_i(x_j^T).
\end{equation}

To mitigate the effects of overconfident predictions arising from domain shift and model uncertainty, we apply label smoothing to the ensemble pseudo-labels before KD:
\begin{equation}
    \label{eq:label_smoothing}
    \tilde{y}_j^T(c) = (1 - \epsilon) \cdot \bar{p}_c(x_j^T) + \frac{\epsilon}{C}, \quad \forall c \in \{1, \dots, C\},
\end{equation}
where $\epsilon \in [0,1]$ is the smoothing factor and $C$ is the number of classes. This transformation shifts the original ensemble distribution $\bar{p}(x_j^T)$ toward the uniform distribution $u(c) = \frac{1}{C}$:
\begin{equation}
    \label{eq:smoothing}
    \tilde{y}_j^T = (1 - \epsilon)\bar{p}(x_j^T) + \epsilon \cdot u.
\end{equation}

As a result, $\tilde{y}_j^T$ satisfies $\tilde{y}_j^T(c) > 0$ for all $c$, ensuring full support over the label space and preventing degenerate cases where a single class dominates.
This smoothing acts as an implicit regularizer by limiting the influence of highly confident predictions from the ensemble.
By reducing the sharpness of the distribution, label smoothing softens the supervision signal received by the student model. This discourages the model from placing excessive confidence in potentially incorrect class predictions, thereby mitigating overfitting to noisy or uncertain targets. 

Once the smoothed pseudo-labels $\tilde{y}_j^T$ are obtained, the student model $f_s$ is trained to align its predictions with these targets. Let $q_j = f_s(x_j^T)$ denote the predicted probability distribution of the student model for input $x_j^T$. We define the SKD loss using the KL divergence between the smoothed teacher signal $\tilde{y}_j^T$ and the student output $q_j$ as:
\begin{equation}
    \label{eq:skd}
    \mathcal{L}_{\text{SKD}} = \frac{1}{N_T} \sum_{j=1}^{N_T} \sum_{c=1}^{C} \tilde{y}_j^T(c) \left( \log \tilde{y}_j^T(c) - \log q_j(c) \right).
\end{equation}

This loss encourages the student model to match the full structure of the softened target distribution, rather than fitting to hard labels. By distributing supervision across all classes in a controlled manner, the KL divergence loss enables stable and effective adaptation to the target domain.

\begin{figure*}[t]
\centering
\includegraphics[width=\textwidth]{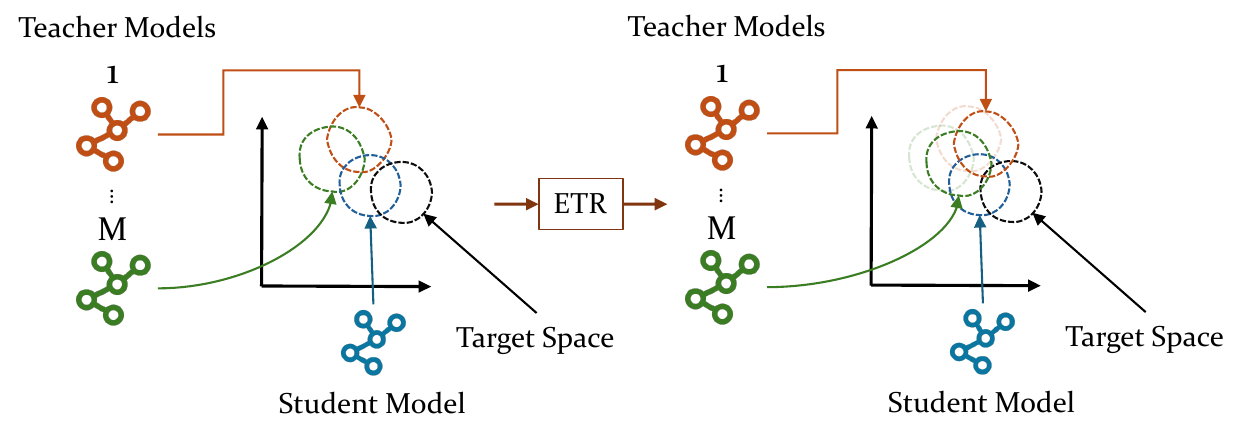}
\caption{Illustration of the ETR process. Each source model initially generates pseudo-labels that reflect its own source-domain distribution, while the student, trained on target-domain pseudo-labels using SKD, represents a distribution closer to the true target domain. Through ETR, teachers are gradually pulled toward the student, which aligns their distributions more closely with the target and enables them to produce more target-aware pseudo-labels.}
\label{fig:ETR}
\end{figure*}

\subsection{Teacher Refinement with EMA}
\label{sec:etr}
After each iteration of adapting the student model $f_s$ to the target domain through SKD, we refine the source models on the server using an EMA of the student’s parameters. This creates a co-adaptive update process in which the teachers help guide the student through their predictions, and the student in turn updates the teachers through parameter interpolation. Figure \ref{fig:ETR} illustrates this bidirectional and mutually refining process, which gradually pulls the teachers, who encode source-domain knowledge, toward the student, which captures target-domain knowledge. This alignment helps the teachers produce more stable and less noisy pseudo-labels for the unlabeled target data in subsequent iterations. The entire process is performed on the server, requires no additional client communication, and the refined teachers are used only during training to improve supervision. After adaptation, only the student is retained for inference.

Let $\theta_s$ denote the corresponding parameters of the student model. The refined parameters $\theta_i'$ for source model $i$ are computed using the following EMA update rule:
\begin{equation}
    \theta_i' \leftarrow (1 - \alpha) \cdot \theta_i + \alpha \cdot \theta_s,
    \label{eq:ema_update_geom}
\end{equation}
where $\alpha \in [0, 1)$ is a momentum coefficient that controls the influence of the previous source parameters versus the student parameters.

{\color{black}

This update rule performs a linear interpolation between the source-specific parameters $\theta_i$ and the target-adapted student parameters $\theta_s$, moving $\theta_i$ along the straight line connecting them in parameter space. Let $h_i, h_s \in \mathcal{H}$ denote the hypotheses induced by $\theta_i$ and $\theta_s$, respectively. Because the student $h_s$ is optimized using target-domain pseudo-labels through SKD, it encodes target-aware information and serves as an implicit anchor for alignment in hypothesis space. Under standard smoothness assumptions \cite{xu_robustness_2012}, this parameter interpolation induces a contraction in hypothesis space, such that:
\begin{equation}
d_{\mathcal{H}}(h_i',\,h_s)
\;\le\;
(1-\alpha) \, d_{\mathcal{H}}(h_i,\,h_s),
\end{equation}
where $h_i'$ denotes the hypothesis induced by the refined parameters $\theta_i'$ and $d_{\mathcal{H}}(\cdot,\cdot)$ is a discrepancy measure over the hypothesis space $\mathcal{H}$.

According to the classical domain adaptation theory \cite{ben-david_analysis_2006, ben-david_theory_2010}, the target risk of a hypothesis $h \in \mathcal{H}$ satisfies an upper bound of the form:
\begin{equation}
\epsilon_T(h)
\;\le\;
\epsilon_S(h)
\;+\;
\frac{1}{2}\, d_{\mathcal{H}\Delta\mathcal{H}}(\mathcal{D}^S,\mathcal{D}^T)
\;+\;
\eta,
\label{eq:uda_bound}
\end{equation}
where $d_{\mathcal{H}\Delta\mathcal{H}}$ measures the maximal disagreement between pairs of hypotheses across the source and target domains. In the FUDA setting, this discrepancy term is driven by disagreement among source hypotheses when evaluated on target samples.

By iteratively updating $\theta_i$ toward the target-adapted student parameters $\theta_s$, EMA-based refinement reduces prediction disagreement on target data between each teacher and the student, i.e.,
\begin{equation}
\mathbb{E}_{x \sim \mathcal{D}^T}
\big[
d\big(p_i'(x),\,p_s(x)\big)
\big]
\;\le\;
\mathbb{E}_{x \sim \mathcal{D}^T}
\big[
d\big(p_i(x),\,p_s(x)\big)
\big]
\end{equation}
where $p_i'(x)$ denotes the predictive distribution induced by the refined parameters $\theta_i'$. Also, because all teachers are jointly contracted toward the same target-aware student hypothesis, their mutual (inter-source) disagreement is also reduced. This process implicitly minimizes an inter-source hypothesis divergence weighted by target-domain signals encoded in the student.

Under standard smoothness assumptions \cite{xu_robustness_2012}, the contraction in parameter space induced by EMA translates into a contraction in hypothesis discrepancy space. This, in turn, reduces the discrepancy between hypotheses on the target domain. Consequently, EMA-based teacher refinement implicitly minimizes a target-aware surrogate of $\mathcal{H}\Delta\mathcal{H}$ divergence within the teacher hypothesis family, thereby tightening the discrepancy term in the target risk bound (Eq. \ref{eq:uda_bound}) for the hypotheses involved in adaptation.
}

To formalize this intuition, the EMA update can be interpreted as minimizing the following objective:
\begin{equation}
    \mathcal{L}_{\text{EMA}}(\theta) = \frac{\lambda}{2} \|\theta - \theta_s\|^2 + \frac{1}{2} \|\theta - \theta_i\|^2,
    \label{eq:ema_regularized_loss}
\end{equation}
whose gradient with respect to $\theta$ is given by:
\begin{equation}
    \nabla \mathcal{L}_{\text{EMA}}(\theta) = \lambda (\theta - \theta_s) + (\theta - \theta_i).
\end{equation}

Setting $\nabla \mathcal{L}_{\text{EMA}} = 0$ yields the optimal solution:
\begin{equation}
    \theta_i' = \frac{\lambda}{1 + \lambda} \theta_s + \frac{1}{1 + \lambda} \theta_i,
\end{equation}
which corresponds exactly to the EMA update in Eq.~\ref{eq:ema_update_geom} when $\lambda = \alpha / (1 - \alpha)$. This confirms that the EMA step is the closed-form minimizer of a weighted proximity objective. 

The overall training process of the proposed SKD-ETR framework is summarized in Algorithm~\ref{alg:skd-etr}.

\begin{algorithm}[htpb]
\caption{SKD-ETR}
\label{alg:skd-etr}
\KwIn{%
    Source domains $\{\mathcal{D}_i^S\}_{i=1}^M$ with labeled data;\\
    Unlabeled target domain $\mathcal{D}^T = \{x_j^T\}_{j=1}^{N_T}$;\\
}
\KwOut{%
    Adapted student model $f_s$;
}

\textbf{Client-side training:}\\
\ForEach{client $i = 1$ to $M$}{
    Train source model $f_i$ on $\mathcal{D}_i^S$ using cross-entropy loss (Eq. \ref{eq:cross_ent})\;
    Send trained parameters $\theta_i = \{\psi_i, \phi_i\}$ to server\;
}

\textbf{Server-side adaptation:}\\
Initialize student model $f_s$ with random parameters $\theta_s$\;

\For{number of epochs}{
    \ForEach{target sample $x_j^T$ in $\mathcal{D}^T$}{
        \ForEach{source model $f_i$}{
            Compute softmax probabilities $p_i(x_j^T)$ (Eq. \ref{eq:client_softmax})\;
        }
        Compute ensemble prediction $\bar{p}(x_j^T)$ (Eq. \ref{eq:ensemble_pred})\;
        Apply label smoothing and compute $\tilde{y}_j^T$ (Eq. \ref{eq:label_smoothing})\;
    }

    Train $f_s$ on $\mathcal{D}^T$ using KL divergence loss with smoothed pseudo labels (Eq. \ref{eq:skd})\;

    \textbf{EMA-based Teacher Refinement:}\\
    \ForEach{source model $f_i$}{
        Update source model $i$ weights via EMA (Eq. \ref{eq:ema_update_geom}) as part of the co-adaptive update process so it can serve again as a teacher in the next iteration\;
    }
}

\Return{$f_s$}\;
\end{algorithm}

\subsubsection{Effect on Decision Boundaries}
In deep classification models, changes in the weight space induce geometric shifts in the decision boundary. Suppose $\theta_s$ and $\theta_i$ define decision boundaries $\mathcal{B}_s$ and $\mathcal{B}_i$, respectively. The EMA update results in a new parameter $\theta_i'$ that lies closer to $\theta_s$, and thus defines a new boundary $\mathcal{B}_i'$ that is a softened version of $\mathcal{B}_i$ toward $\mathcal{B}_s$.
If the student model has learned a flatter, more domain-invariant boundary due to its exposure to multi-teacher soft supervision on the target domain, then moving $\theta_i$ in its direction reduces the sharpness and domain specificity of $\mathcal{B}_i$. This results in smoother transitions between classes and better generalization to unseen samples from the target distribution.

\section{Experimental settings}
\label{sec:implementation_details}

\subsection{Datasets}
\label{sec:datasets}
\begin{figure}[htbp]
\centering
\includegraphics[width=1\textwidth]{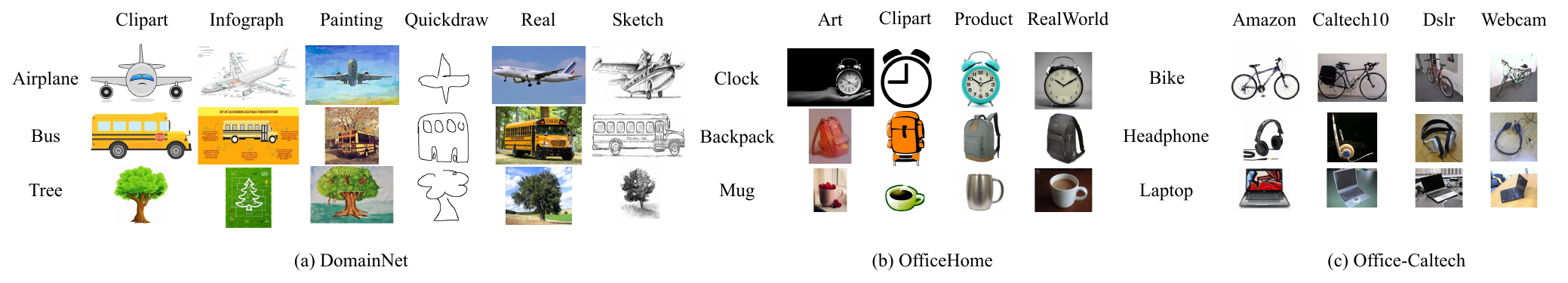}
\caption{Sample images from different domains in different datasets. (a) DomainNet, (b) OfficeHome, and (c) Office-Caltech. Each row represents a category, and each column corresponds to a visual domain within that dataset.}
\label{fig:samples}
\end{figure}

We evaluate the effectiveness of our proposed method on three widely used multi-domain benchmarks for unsupervised domain adaptation.
\textbf{OfficeHome}~\cite{venkateswara_deep_2017} contains approximately 15,500 images categorized into 65 classes spanning four distinct domains: Art, Clipart, Product, and Real-World. This dataset presents significant domain shift, making it a standard and challenging benchmark for domain adaptation. \textbf{Office-Caltech}~\cite{gong_geodesic_2012} combines the Office-31 \cite{hutchison_adapting_2010} dataset with the Caltech-256 \footnote{https://data.caltech.edu/records/20087} dataset, resulting in a unified benchmark with 2,533 images from four domains: Amazon, Webcam, DSLR, and Caltech, restricted to 10 common object categories. \textbf{DomainNet}~\cite{peng_moment_2019} is the largest and most diverse benchmark in our evaluation, comprising 48,129 images across six highly diverse domains: Clipart, Real, Sketch, Infograph, Painting, and Quickdraw, with a shared label space of 345 categories.
These datasets collectively cover a wide range of domain shifts, image styles, and semantic spaces, allowing for comprehensive evaluation of adaptation performance under both mild and severe distribution shifts. Figure~\ref{fig:samples} illustrates sample images from different domains and classes across the three benchmarks.

\subsection{Implementation Details}
All local client models are trained using stochastic gradient descent with a momentum of 0.9, a learning rate of 3e-2, and a batch size of 32. We apply a 5\% linear warm-up schedule at the beginning of training. Each client trains its local model for 20 epochs. The student model on the server is trained for 10 epochs using the SKD procedure described in Section~\ref{sec:skd}.

The smoothing factor $\epsilon$ used in generating soft pseudo-labels is set to 0.5, as discussed in Section~\ref{sec:sa_epsilon}, which balances confidence suppression with class diversity in the supervision signal. For teacher refinement, the EMA momentum coefficient $\alpha$ is set to 0.15 based on the analysis in Section~\ref{sec:sa_alpha}, encouraging gradual alignment between the source models and the student model while retaining local domain-specific knowledge.

We follow the leave-one-domain-out protocol \cite{peng_federated_2019, hong_federated_2021, feng_kd3a_2021, xiao_unsupervised_2024, abedi_one-shot_2025, yi_fdac_2025}, where in each experiment, one domain is considered as the unlabeled target and the remaining domains as labeled sources. For augmentation, we apply random cropping and random horizontal flipping, and normalize all images using the ImageNet \cite{deng_imagenet_2009} mean and standard deviation.
All experiments are conducted using the same backbone architecture and training setup across datasets to ensure comparability and reproducibility. For the bottleneck and head architecture, we follow the design introduced in \cite{abedi_euda_2024}, which provides a consistent and well-established configuration for feature compression and adaptation.

\subsection{Baselines and Evaluation Metric}
We compare our method against three categories of baselines. MSUDA methods include SSD~\cite{li_multidomain_2024}, SImpAl~\cite{venkat_your_2020}, and MDAMA~\cite{dayal_leveraging_2025}. MSFDA methods include TransDA~\cite{yang_transformer-based_2021}, SHOT~\cite{liang_we_2020}, and DECISION~\cite{ahmed_unsupervised_2021}. FUDA methods include FedSD2C \cite{zhang_one-shot_2024}, FedPFT \cite{beitollahi_parametric_2024}, FADA~\cite{peng_federated_2019}, FADE~\cite{hong_federated_2021}, KD3A~\cite{feng_kd3a_2021}, FDKD~\cite{xiao_unsupervised_2024}, SEA+MSPL~\cite{abedi_one-shot_2025}, and FDAC~\cite{yi_fdac_2025}. Among these methods, FedSD2C~\cite{zhang_one-shot_2024} and FedPFT~\cite{beitollahi_parametric_2024} are one-shot FL approaches that we extend to the FUDA setting, while SEA+MSPL~\cite{abedi_one-shot_2025} is a one-shot FUDA method.

We evaluate all methods on OfficeHome, Office-Caltech, and DomainNet using the standard protocol in FUDA literature, where each domain is treated as the target once and the remaining domains serve as sources. Each source domain is assigned to a separate client, resulting in $M = L - 1$ clients per experiment, where $L$ is the total number of domains. The evaluation metric is top-1 classification accuracy on the unlabeled target domain. We report per-domain results as well as the average accuracy across all domains.

\section{Results and Analysis}
\label{sec:results}
\subsection{Results}
Tables~\ref{table:comp_oh} and \ref{table:comp_dn} compare the performance of our proposed SKD-ETR method with state-of-the-art baselines on the OfficeHome, DomainNet, and Office-Caltech datasets. SKD-ETR consistently achieves the highest average accuracy on OfficeHome and DomainNet, demonstrating its effectiveness in handling diverse and complex domain shifts. On Office-Caltech, it delivers performance comparable to the best-performing baseline while significantly reducing the number of trainable parameters, communication overhead, and the size of sharable model weights. 

\begin{figure}[htbp]
\centering
\includegraphics[width=\textwidth]{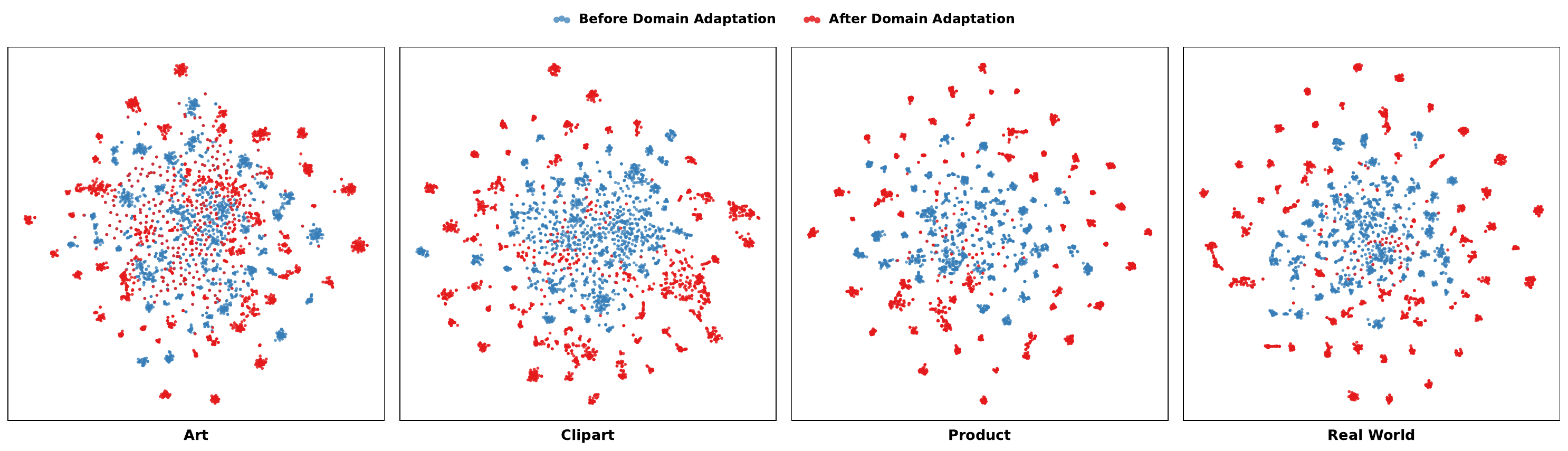}
\caption{t-SNE visualizations of the learned feature representations of the student model on the OfficeHome dataset before and after domain adaptation.}
\label{fig:tsne}
\end{figure}

Additionally, Figure~\ref{fig:tsne} presents the representations learned by the student model using t-SNE before and after domain adaptation with SKD-ETR. The visualization clearly shows that our approach leads to better feature alignment and class separability in the target domain. This highlights the effect of adaptation on the student model. To further quantify these representational shifts, we analyzed the movement of the decision boundary using Flip Rate and KL Divergence between the student model before and after adaptation. The Flip Rate measures the fraction of samples whose predicted labels change after adaptation, while the KL divergence quantifies the average shift in prediction distributions. As shown in Table~\ref{table:decision_boundary_shift}, all domains exhibit high Flip Rate and KL divergence values, indicating substantial prediction and distributional changes following training. These results confirm that SKD-ETR effectively adapts the decision boundaries toward the target domain, resulting in improved alignment and generalization.

\begin{table}[htbp]
\caption{Quantitative analysis of decision boundary shifts before and after adaptation on the OfficeHome dataset using Flip Rate and KL Divergence.}
\label{table:decision_boundary_shift}
\centering
\resizebox{0.68\textwidth}{!}{
\begin{tabular}{lcc}
\toprule
\textbf{Domain} & \textbf{Flip Rate (↑)} & \textbf{KL Divergence (↑)} \\
\midrule
Art & 0.9749 & 17.4327 \\
Clipart & 0.9835 & 15.8056 \\
Product & 0.9905 & 16.2890 \\
Real World & 0.9837 & 16.2571 \\
\bottomrule
\end{tabular}
}
\end{table}

\begin{table*}[htbp]
\caption{Comparison of accuracy (\%) on OfficeHome and Office-Caltech benchmarks across MSUDA, MSFDA, and FUDA methods. $\dagger$ denotes one-shot FL.}
\label{table:comp_oh}
\centering
\def\arraystretch{1.2}
\resizebox{\textwidth}{!}{
\begin{tabular}{c|c|cccc|c|cccc|c}
\toprule
 & \multirow{2}{*}{Model} & \multicolumn{5}{c|}{OfficeHome} & \multicolumn{5}{c}{Office-Caltech} \\ \cline{3-12}
 & & $\rightarrow$A & $\rightarrow$C & $\rightarrow$P & $\rightarrow$R & Avg. & $\rightarrow$A & $\rightarrow$C & $\rightarrow$D & $\rightarrow$W  & Avg. \\ \midrule
\multirow{3}{*}{\textbf{MSUDA}} 
& SImpAl \cite{venkat_your_2020} & 73.4 & 62.4 & 81.0 & 82.7 & 74.8 & 95.6 & 94.6 & 100.0 & 100.0 & 97.5 \\
& SSD \cite{li_multidomain_2024} & 72.5 & 64.5 & 81.2 & 83.2 & 75.4 & 95.8 & 95.0 & 100.0 & 99.1 & 97.5 \\
& MDAMA \cite{dayal_leveraging_2025} & 77.7 & 69.9 & 85.6 & 87.1 & 80.0 & 96.9 & 96.7 & 100.0 & 100.0 & 98.4 \\
\midrule
\multirow{3}{*}{\textbf{MSFDA}} 
& SHOT \cite{liang_we_2020} & 82.9 & 82.8 & 59.3 & 72.2 & 74.3 & 96.4 & 96.2 & 98.5 & 99.7 & 97.7 \\
& DECISION \cite{ahmed_unsupervised_2021} & 83.6 & 84.4 & 59.4 & 74.5 & 75.5 & 95.9 & 95.9 & 99.7 & 99.6 & 98.4 \\
& TransDA \cite{yang_transformer-based_2021} & 76.4 & 56.5 & 84.4 & 86.4 & 75.9 & 96.4 & 97.4 & 100.0 & 98.8 & 98.2 \\
\midrule
\multirow{8}{*}{\textbf{FUDA}} 
& FedSD2C$^{\dagger}$ \cite{zhang_one-shot_2024} & 24.8 & 22.4 & 35.1 & 32.8 & 28.8 & 59.2 & 50.1 & 78.3 & 93.2 & 70.2 \\
& FADA \cite{peng_federated_2019} & 67.4 & 68.1 & 48.6 & 54.0 & 59.5 & 82.4 & 88.7 & 87.1 & 88.1 & 87.1 \\
& FedSD2C$^{\dagger}$ \cite{zhang_one-shot_2024} & 24.8 & 22.4 & 35.1 & 32.8 & 28.8 & 59.2 & 50.1 & 78.3 & 93.2 & 70.2 \\
& FADE \cite{hong_federated_2021} & 76.9 & 61.2 & 85.3 & 84.6 & 77.0 & 96.5 & 95.9 & 99.8 & 99.6 & 98.2 \\
& FDKD \cite{xiao_unsupervised_2024} & 85.3 & 84.1 & 63.0 & 75.6 & 77.0 & 98.7 & 97.6 & 99.5 & 99.5 & \textbf{99.0} \\
& KD3A \cite{feng_kd3a_2021} & 79.2 & 62.3 & 87.5 & 87.3 & 79.1 & 96.5 & 95.3 & 97.8 & 99.1 & 97.2 \\
& FDAC \cite{yi_fdac_2025} & 80.2 & 63.3 & 89.2 & 88.6 & 80.3 & 96.9 & 97.0 & 100.0 & 100.0 & 98.5 \\
& SEA + MSPL$^{\dagger}$ \cite{abedi_one-shot_2025} & 83.2 & 76.6 & 91.2 & 90.8 & 85.4 & 96.1 & 97.2 & 99.4 & 99.7 & 98.1 \\
& SKD-ETR (Ours)$^{\dagger}$ & $84.1 \pm 0.1$ & $76.2 \pm 0.1$ & $91.3 \pm 0.1$ & $92.1 \pm 0.2$ & $\textbf{85.9} \pm \textbf{0.1}$ & $97.7 \pm 0.1$ & $96.8 \pm 0.1$ & $100.0 \pm 0.0$ & $100.0 \pm 0.0$ & $98.6 \pm 0.1$ \\ \bottomrule
\end{tabular}%
}
\end{table*}

\begin{table*}[htbp]
\caption{Comparison of accuracy (\%) on DomainNet benchmarks across MSUDA, MSFDA, and FUDA methods. $\dagger$ denotes one-shot FL.}
\label{table:comp_dn}
\centering
\def\arraystretch{1.2}
\resizebox{0.75\textwidth}{!}{
\begin{tabular}{c|c|cccccc|c}
\toprule
 & {Model} & $\rightarrow$C & $\rightarrow$I & $\rightarrow$P & $\rightarrow$Q & $\rightarrow$R & $\rightarrow$S & Avg. \\ \midrule
\multirow{3}{*}{\textbf{MSUDA}} 
& SSD \cite{li_multidomain_2024} & 67.2 & 21.7 & 52.4 & 20.8 & 67.8 & 55.3 & 47.5 \\
& SImpAl \cite{venkat_your_2020} & 66.4 & 26.5 & 56.6 & 18.9 & 68.0 & 55.5 & 48.6 \\
& MDAMA \cite{dayal_leveraging_2025} & 69.2 & 26.5 & 57.7 & 19.6 & 69.7 & 56.3 & 49.8 \\
\midrule
\multirow{3}{*}{\textbf{MSFDA}} 
& TransDA \cite{yang_transformer-based_2021} & 64.9 & 22.7 & 54.1 & 11.8 & 60.8 & 49.6 & 44.0 \\
& SHOT \cite{liang_we_2020} & 61.7 & 22.2 & 52.6 & 12.2 & 67.7 & 48.6 & 44.2 \\
& DECISION \cite{ahmed_unsupervised_2021} & 61.5 & 21.6 & 54.6 & 18.9 & 67.5 & 51.0 & 45.9 \\
\midrule
\multirow{8}{*}{\textbf{FUDA}} 
& FedSD2C \cite{zhang_one-shot_2024}$^{\dagger}$ & 30.8 & 7.1 & 20.6 & 18.7 & 32.6 & 18.0 & 21.3 \\
& FedPFT \cite{beitollahi_parametric_2024}$^{\dagger}$ & 28.8 & 12.1 & 34.9 & 9.0 & 58.4 & 20.1 & 27.2 \\
& FADA \cite{peng_federated_2019} & 45.3 & 16.3 & 38.9 & 7.9 & 46.7 & 26.8 & 30.3 \\
& FADE \cite{hong_federated_2021} & 68.9 & 27.4 & 57.8 & 12.3 & 72.4 & 54.4 & 48.9 \\
& KD3A \cite{feng_kd3a_2021} & 71.3 & 28.5 & 60.7 & 15.8 & 73.4 & 58.7 & 51.7 \\
& FDKD \cite{xiao_unsupervised_2024} & 73.5 & 28.6 & 63.6 & 18.9 & 68.8 & 61.9 & 52.5 \\
& SEA + MSPL \cite{abedi_one-shot_2025}$^{\dagger}$ & 69.7 & 33.5 & 60.9 & 19.7 & 74.9 & 60.8 & 53.2 \\
& FDAC \cite{yi_fdac_2025} & 74.0 & 30.0 & 62.4 & 19.0 & 75.3 & 62.0 & 53.8 \\
& SKD-ETR (Ours)$^{\dagger}$ & $74.2 \pm 1.1$ & $34.5 \pm 1.4$ & $64.1 \pm 0.7$ & $10.8 \pm 0.3$ & $78.3 \pm 1.0$ & $64.5 \pm 0.2$ & $\textbf{54.4} \pm \textbf{0.8}$ \\ \bottomrule
\end{tabular}%
}
\end{table*}

\subsection{Ablation Study}

This section presents a comprehensive ablation study to evaluate the contribution of each component within the proposed framework. We begin by analyzing the effects of the main components, SKD and ETR, and subsequently examine different student model initialization strategies, pseudo-labeling techniques, and backbone architectures.

\subsubsection{Effect of Different Components of the Model}
We assess the contribution of each component in SKD-ETR by removing ETR, SKD, and both. As shown in Table~\ref{table:ablation}, removing either component reduces performance, while removing both results in the lowest accuracy. This confirms that both SKD and ETR are critical to the overall effectiveness of our method.

\begin{table}[htbp]
\caption{Ablation results showing the impact of removing SKD and ETR on target domain performance.}
\label{table:ablation}
\centering
\resizebox{0.7\textwidth}{!}{
\begin{tabular}{l|cccc|c}
\toprule
{Model} & $\rightarrow$A & $\rightarrow$C & $\rightarrow$P & $\rightarrow$R & Avg. \\ \midrule
Ours & 84.1 & 76.2 & 91.3 & 92.1 & \textbf{85.9} \\
Ours w/o ETR & 83.3 & 73.0 & 91.2 & 91.9 & 84.7 \\
Ours w/o SKD & 81.7 & 72.4 & 90.1 & 90.8 & 83.7 \\
Ours w/o ETR \& SKD, w/ KL Divergence & 80.9 & 71.6 & 89.7 & 90.0 & 83.0 \\
\bottomrule
\end{tabular}
}
\end{table}

\subsubsection{Effect of Different Student Model Initializations}
Table~\ref{table:student_init} compares several strategies for initializing the student model, including random initialization, averaged-teacher initialization, lowest-entropy source, and random-source initialization.
In random initialization, the student starts with entirely new parameters, independent of any source model.
The averaged-teacher initialization computes the mean of all source teachers’ bottleneck and head parameters to provide a unified starting point.
The lowest-entropy source initialization selects the source model with the most confident predictions, i.e., the source model yielding the lowest output entropy, as the student’s initialization.
Additionally, the random-source initialization simply chooses one source model at random for initialization.
Based on the results, random initialization achieves the highest average performance across the four OfficeHome domains. Transferring parameters from source models, either individually or through averaging, often introduces domain-specific biases or noisy supervision that impede adaptation. In contrast, random initialization provides an unbiased starting point, allowing SKD to guide the student toward more domain-invariant representations. We therefore adopt this initialization strategy in all subsequent experiments.

\begin{table}[htbp]
\caption{Ablation study of our method with different student model initialization strategies.}
\label{table:student_init}
\centering
\resizebox{0.6\textwidth}{!}{
\begin{tabular}{l|cccc|c}
\toprule
{Initialization Strategy} & $\rightarrow$A & $\rightarrow$C & $\rightarrow$P & $\rightarrow$R & Avg. \\ \midrule
Random initialization (Ours) & 84.1 & 76.2 & 91.3 & 92.1 & \textbf{85.9} \\
Averaging & 82.8 & 74.0 & 90.7 & 91.7 & 84.8 \\
Lowest-entropy source model & 83.4 & 73.4 & 88.1 & 90.8 & 83.9 \\
Random source model & 83.6 & 71.9 & 88.6 & 90.7 & 83.7 \\
\bottomrule
\end{tabular}
}
\end{table}

\subsubsection{Effect of Pseudo Labeling Strategies}
Table~\ref{table:pseduo_labeling} reports the performance of the model using different pseudo-labeling strategies. Confidence Thresholding~\cite{saito_asymmetric_2017} discards pseudo-labels whose maximum confidence is below a fixed threshold (0.8 in our experiments), which helps remove noisy labels but also excludes a large portion of target data. Confidence Weighting~\cite{zou_confidence_2020} assigns weights to each pseudo-label based on its maximum confidence, while Entropy weighting~\cite{long_conditional_2018} uses the entropy of predictions to weight supervision, assigning higher weight to more confident outputs. 
While these approaches alleviate the effect of uncertain labels, they frequently introduce bias toward overconfident predictions or cause considerable data loss. 
In contrast, our proposed SKD method retains all target samples and regularizes supervision through label smoothing. This prevents overconfidence and yields more stable pseudo-labels. Consequently, SKD achieves the highest average performance across all domains.

\begin{table}[htbp]
\caption{Performance comparison of our model using different pseudo labeling strategies.}
\label{table:pseduo_labeling}
\centering
\resizebox{0.6\textwidth}{!}{
\begin{tabular}{l|cccc|c}
\toprule
{Pseudo Labeling Strategy} & $\rightarrow$A & $\rightarrow$C & $\rightarrow$P & $\rightarrow$R & Avg. \\ \midrule
SKD (Ours) & 84.1 & 76.2 & 91.3 & 92.1 & \textbf{85.9} \\
Entropy Weighting \cite{long_conditional_2018} & 83.8 & 73.9 & 89.7 & 89.1 & 84.1 \\
Confidence Weighting \cite{zou_confidence_2020} & 83.8 & 72.6 & 87.5 & 90.3 & 83.6 \\
Confidence Thresholding \cite{saito_asymmetric_2017} & 76.4 & 62.6 & 85.1 & 85.8 & 77.5 \\
\bottomrule
\end{tabular}
}
\end{table}

\subsubsection{Effect of Different Backbones}
\label{sec:sa_backbone}
We evaluate the effect of different backbone architectures, including ResNet-50, ResNet-101, ViT, and DINOv2, on our framework by comparing SKD-ETR with a baseline variant trained only with KL divergence, without SKD and ETR. As shown in Table~\ref{table:backbones}, SKD-ETR improves performance across all backbones, yielding modest yet consistent gains on convolutional networks and more pronounced improvements on transformer-based models. All backbones are kept frozen in our method. In particular, DINOv2 combined with SKD-ETR achieves the highest average performance. This demonstrates that our method is architecture-agnostic, benefits from stronger representation capacity, and remains effective for both convolutional and transformer-based backbones.

\begin{table}[htbp]
\caption{Performance comparison of our model across different backbones.}
\label{table:backbones}
\centering
\def\arraystretch{1.7}
\resizebox{0.8 \textwidth}{!}{
\begin{tabular}{l|c|cccc|c}
\toprule
Backbone & Method & $\rightarrow$A & $\rightarrow$C & $\rightarrow$P & $\rightarrow$R & Avg. \\ \midrule
\multirow{2}{*}{Resnet-50 \cite{he_deep_2016}} &
                                            w/o ETR \& SKD, w/ KL Divergence & 64.6 & 45.8 & 74.2 & 78.2 & 65.7 \\
                                            & SKD-ETR & 66.0 & 47.2 & 77.4 & 80.3 & 67.7  \\ \hline

\multirow{2}{*}{Resnet-101 \cite{he_deep_2016}} &
                                            w/o ETR \& SKD, w/ KL Divergence & 67.4 & 51.9 & 77.8 & 80.6 & 69.4 \\
                                            & SKD-ETR & 71.4 & 52.6 & 78.9 & 80.9 & 71.0 \\ \hline

\multirow{2}{*}{ViT \cite{dosovitskiy_image_2021}} &
                                            w/o ETR \& SKD, w/ KL Divergence & 66.9 & 56.2 & 78.6 & 80.5 & 70.5 \\
                                            & SKD-ETR & 69.9 & 59.3 & 82.2 & 82.8 & 73.5 \\ \hline

\multirow{2}{*}{DINOv2 \cite{oquab_dinov2_2024}} &
                                            w/o ETR \& SKD, w/ KL Divergence & 80.9 & 71.6 & 89.7 & 90.0 & 83.0 \\
                                            & SKD-ETR & 84.1 & 76.2 & 91.3 & 92.1 & \textbf{85.9} \\
\bottomrule
\end{tabular}
}
\end{table}

\subsection{Sensitivity Analysis}
This section presents a sensitivity analysis of our model and examines how various factors affect its performance. Specifically, we analyze the impact of the smoothing factor $\epsilon$ and the EMA decay factor $\alpha$. Furthermore, we evaluate the robustness of the proposed method to parameter noise to assess its stability and reliability under noisy client conditions.

\subsubsection{Effect of $\epsilon$}
\label{sec:sa_epsilon}
\begin{figure}[htbp]
\centering
\includegraphics[width=0.6\textwidth]{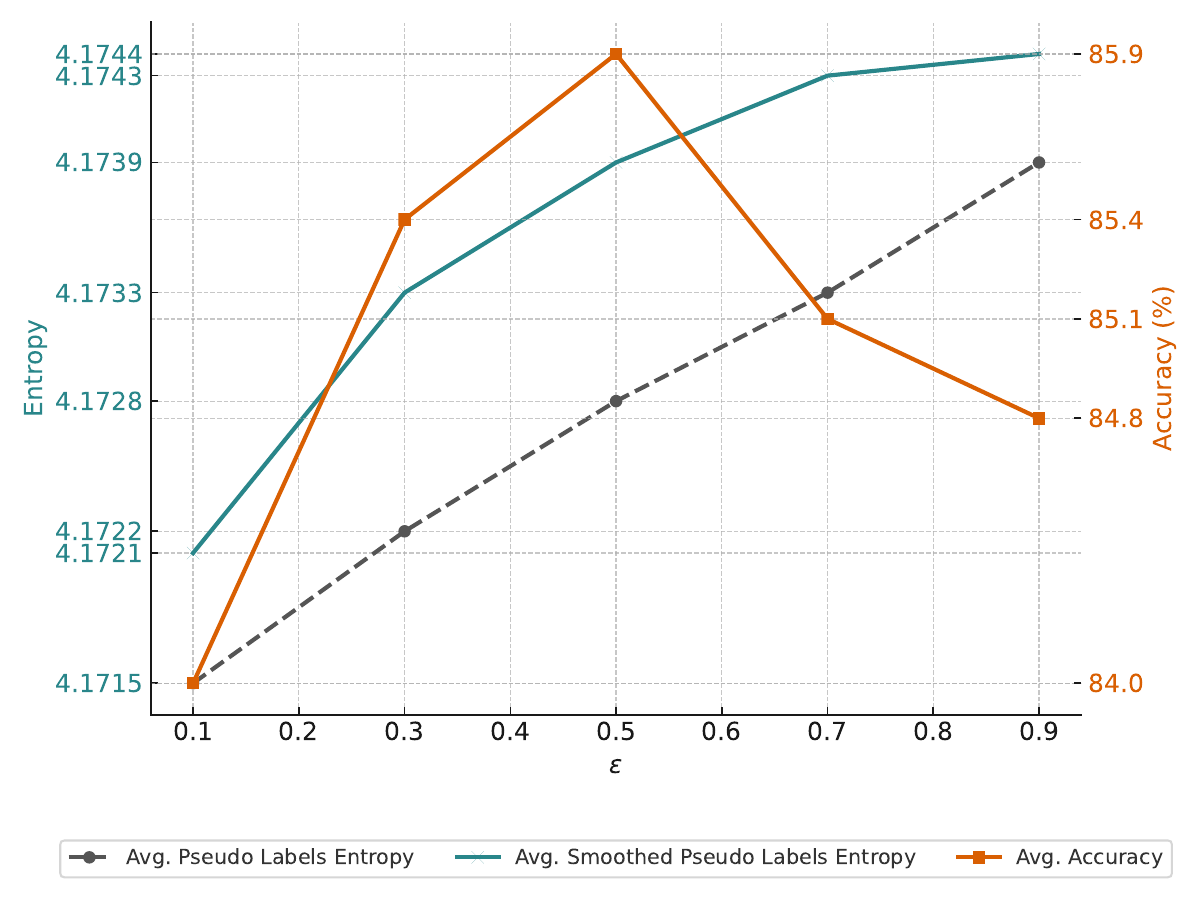}
\caption{Sensitivity analysis of the smoothing factor $\epsilon$ in SKD.}
\label{fig:epsilon_ablation}
\end{figure}

We analyze the impact of the smoothing factor $\epsilon$ in Eq.~\ref{eq:smoothing}. As shown in Figure~\ref{fig:epsilon_ablation}, the best target performance is achieved at $\epsilon = 0.5$, which offers an effective trade-off between preserving informative signals from the source predictions and reducing noise in the pseudo labels. In addition to accuracy, we also visualize the average entropy of the pseudo labels before and after smoothing during training process. As expected, the smoothed pseudo labels consistently exhibit higher entropy, indicating increased uncertainty. This controlled uncertainty is beneficial up to $\epsilon = 0.5$, as it helps regularize noisy predictions. When $\epsilon$ is too low, the pseudo-labels exhibit low entropy and tend to preserve incorrect high-confidence predictions, which can mislead the student during early training. In contrast, very high values of $\epsilon$ produce overly diffuse distributions that may obscure important class distinctions.

{\color{black}In addition, we compare the fixed choice of $\epsilon$ with two adaptive smoothing strategies. First, we consider an entropy-adaptive $\epsilon$, where the smoothing factor is increased for pseudo-labels with higher predictive entropy. This strategy implicitly accounts for heterogeneous uncertainty arising from class imbalance and severe source–target mismatch. Second, we consider a time-decayed schedule, where $\epsilon$ decreases from a high initial value (e.g., $0.9$) to a lower value (e.g., $0.1$) over training.

As shown in Table \ref{table:label_smoothig}, entropy-adaptive smoothing yields slight improvements in some extreme-shift scenarios but generally underperforms the fixed $\epsilon$ setting, as confident but incorrect pseudo-labels are not sufficiently suppressed and can mislead the student during adaptation. The time-decayed $\epsilon$ schedule consistently underperforms the fixed $\epsilon$ setting, as the predefined decay cannot adapt to sample-level or domain-specific uncertainty. These results suggest that a fixed $\epsilon$ provides a stable and robust regularization mechanism under long-tailed and highly divergent target domains, without introducing additional sensitivity or assumptions.}

\begin{table}[htbp]
\caption{Performance comparison on different label smoothing strategies for pseudo-label refinement.}
\label{table:label_smoothig}
\centering
\resizebox{0.6\textwidth}{!}{
\begin{tabular}{l|cccc|c}
\toprule
{$\epsilon$ Strategy} & $\rightarrow$A & $\rightarrow$C & $\rightarrow$P & $\rightarrow$R & Avg. \\ \midrule
Fixed (Ours) & 84.1 & 76.2 & 91.3 & 92.1 & \textbf{85.9} \\
Entropy-adaptive & 83.3 & 76.5 & 91.0 & 91.9 & 85.6 \\
Time-adaptive & 83.4 & 71.8 & 91.2 & 91.3 & 84.4 \\
\bottomrule
\end{tabular}
}
\end{table}

\subsubsection{Effect of $\alpha$}
\label{sec:sa_alpha}
\begin{figure}[htbp]
\centering
\includegraphics[width=0.5\textwidth]{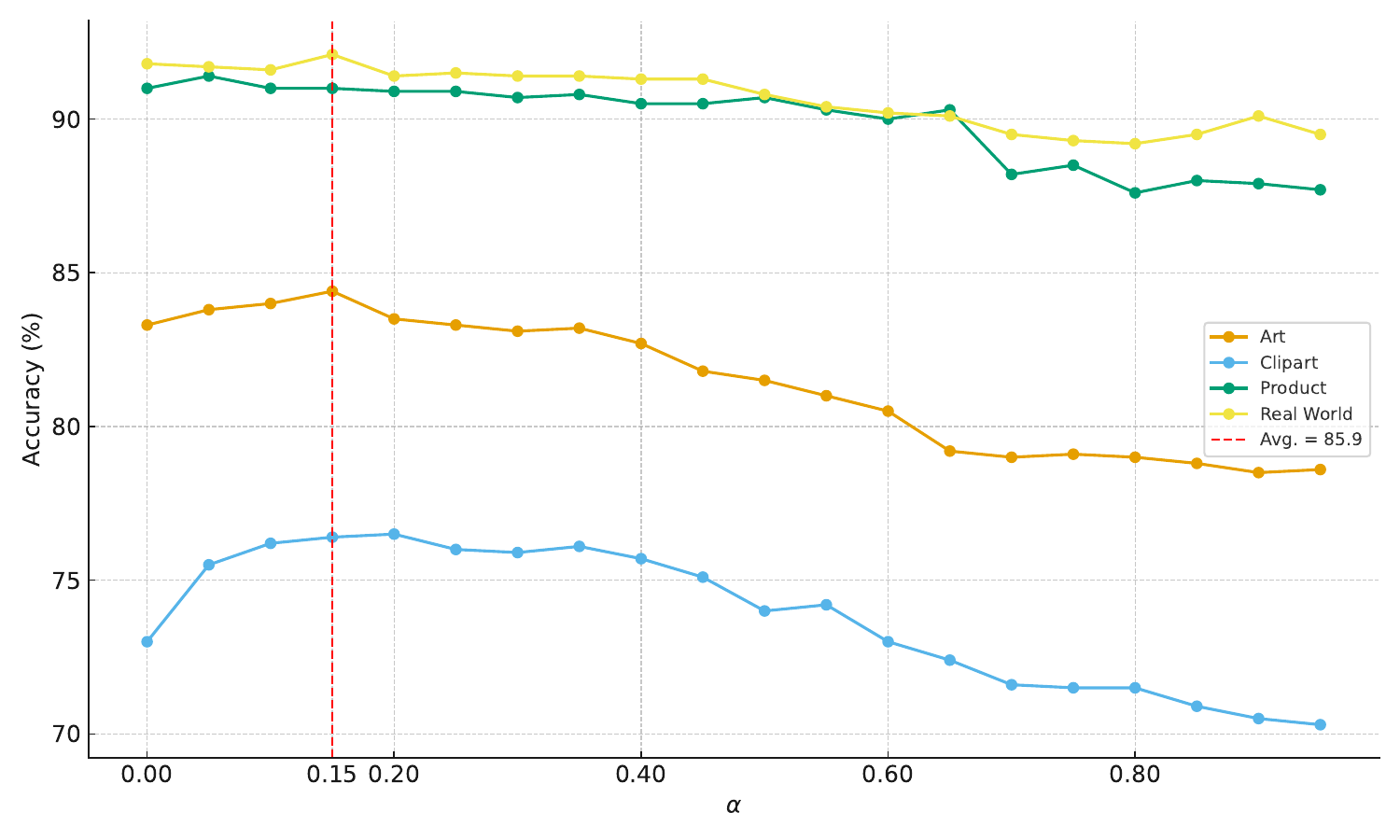}
\caption{Sensitivity analysis of our model across different values of $\alpha$.}
\label{fig:alpha_ablation}
\end{figure}

We conduct a sensitivity analysis on the EMA decay factor $\alpha$ in Eq. \ref{eq:ema_update_geom} to study its effect on model performance. As shown in Figure~\ref{fig:alpha_ablation}, the value of $\alpha$ was varied from $0$ to $0.95$ in increments of $0.05$. The best results are achieved when $\alpha = 0.15$, which strikes a balance between preserving the original source model knowledge and incorporating the generalized student knowledge.

\subsubsection{Robustness to Parameter Noise}
To assess the robustness of SKD-ETR, we introduced Gaussian noise into the clients' parameters to simulate potential privacy-preserving perturbations and the effect of malicious or unstable client updates. As shown in Table~\ref{table:noise_sensitivity}, SKD-ETR is very robust to Gaussian noise injected into source model parameters before transmission. Its performance remains stable with only minor drops for $\sigma = 0.01$ and $\sigma = 0.02$, and it still retains a reasonable level of accuracy even under heavy noise at $\sigma = 0.05$. We also report results for FedAvg~\cite{mcmahan_communication-efficient_2017} under the same setting, which shows much larger degradation. This highlights the strong robustness of SKD-ETR to parameter perturbations.

\begin{table}[htbp]
\caption{Sensitivity analysis of SKD-ETR to Gaussian noise injected into source model parameters before transmission to the server.}
\label{table:noise_sensitivity}
\centering
\def\arraystretch{1.5}
\resizebox{0.7\textwidth}{!}{
\begin{tabular}{c|c|cccc|c}
\toprule
{Gaussian noise $\sigma$} & Method & $\rightarrow$A & $\rightarrow$C & $\rightarrow$P & $\rightarrow$R & Avg. \\ \midrule
\multirow{2}{*}{0 (No Noise)}
                                            & FedAvg & 66.4 & 63.0 & 82.7 & 83.8 & 74.0 \\
                                            & SKD-ETR & 84.1 & 76.2 & 91.3 & 92.1 & \textbf{85.9}  \\ \hline
\multirow{2}{*}{0.01}
                                            & FedAvg & 63.8 & 61.5 & 82.5 & 81.6 & 72.4 (-1.6) \\
                                            & SKD-ETR & 83.8 & 76.1 & 90.8 & 91.5 & 85.6 (-0.3) \\ \hline
\multirow{2}{*}{0.02}
                                            & FedAvg & 49.0 & 53.0 & 63.7 & 81.6 & 61.8 (-12.2) \\
                                            & SKD-ETR & 83.2 & 75.4 & 90.7 & 91.3 & 85.2 (-0.7) \\ \hline
\multirow{2}{*}{0.05}
                                            & FedAvg & 2.3 & 3.0 & 2.3 & 2.1 & 2.4 (-71.6) \\
                                            & SKD-ETR & 25.9 & 25.6 & 37.3 & 42.2 & 32.8 (-53.1) \\    
\bottomrule
\end{tabular}
}
\end{table}

\subsubsection{Effect of ETR on Teacher models}
\begin{figure}[htbp]
\centering
\includegraphics[width=\textwidth]{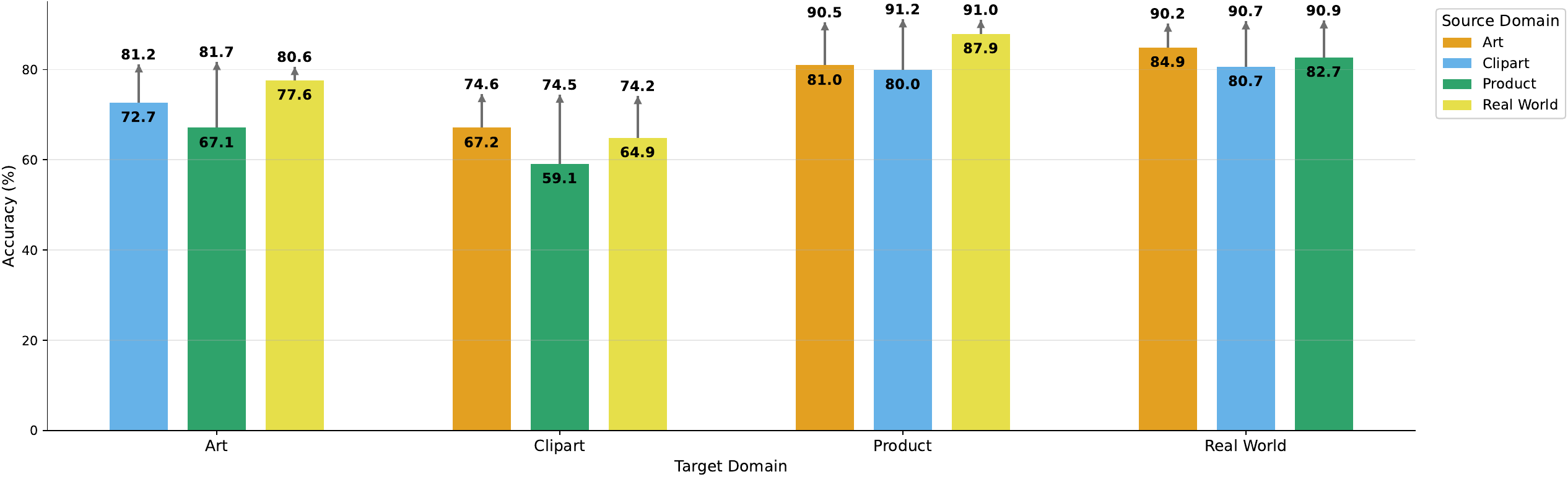}
\caption{Target-domain accuracy of each source model before and after adaptation process.}
\label{fig:teacher_acc}
\end{figure}

To examine the impact of ETR, we evaluate the target-domain performance of each source model before and after the adaptation process, during which multiple rounds of teacher refinement are applied. As shown in Figure~\ref{fig:teacher_acc}, ETR consistently improves the target-domain accuracy of all teacher models, indicating that the refinement process effectively pulls each source hypothesis toward the target domain. For example, when Art is the target domain, the source model trained on Clipart improves from 72.7\% accuracy before adaptation to 81.2\% after the refinement process. Additionally, ETR reduces the pairwise KL divergence between teacher predictions on the target domain (see Table~\ref{table:kl_diversity}). This improvement demonstrates that ETR progressively decreases inter-teacher disagreement and enhances hypothesis alignment prior to pseudo-label aggregation.

Although refined teachers exhibit clear performance gains, their accuracy remains lower than that of the adapted student model. This is expected, as the student is directly optimized on the target-domain, while teachers are updated indirectly via parameter interpolation. These results suggest that ETR primarily serves as a training-time alignment mechanism that improves teacher quality and pseudo-label reliability. While ensembles of refined teachers may offer additional robustness in safety-critical scenarios, the student provides a more favorable accuracy–efficiency trade-off and is therefore retained for inference.

\begin{table}[htbp]
\caption{Average pairwise KL divergence between teacher model predictions on the target domain before and after adaptation.}
\label{table:kl_diversity}
\centering
\resizebox{0.5\textwidth}{!}{
\begin{tabular}{l|cc}
\toprule
\textbf{Target Domain} & \textbf{Before Adaptation} & \textbf{After Adaptation} \\
\midrule
Art        & 2.15 & 0.00044 \\
Clipart   & 2.55 & 0.00044 \\
Product   & 1.45 & 0.00048 \\
Real World & 1.67 & 0.00049 \\
\bottomrule
\end{tabular}
}
\end{table}

\subsection{Computational Overhead and Resource Efficiency}
To analyze the computational cost of SKD-ETR, we measured the training time, GPU memory footprint, floating-point operations (FLOPs), and energy consumption of both teacher and student models on an NVIDIA GeForce GTX 1080 Ti GPU. Results are summarized in Table~\ref{table:compute_analysis}. Each teacher model requires approximately 70.7 GFLOPs per forward–backward pass, 364 MB of parameters, and peaks at 0.81 GB of GPU memory, with training time averaging 1,560 seconds. The student model, which performs the cross-domain adaptation, incurs a moderate increase in memory usage and training time while maintaining the same FLOP count and parameter size. The total estimated energy consumption for training student model remains low at approximately 0.05 kg CO$_2$. The inference time for a batch of 32 samples is 0.01 seconds. It is also important to note that, during adaptation, the teachers remain in non-training mode and contribute only lightweight forward passes. Consequently, both the ensemble pseudo-label generation and the EMA refinement scale linearly with the number of clients $M$. These operations are performed only once during adaptation and do not impact the $O(1)$ inference cost after the student model is trained. These results confirm that SKD-ETR achieves effective adaptation with modest computational and energy requirements.

\begin{table}[htbp]
\caption{Computational overhead and resource usage for SKD-ETR across teacher and student models on the OfficeHome dataset.}
\label{table:compute_analysis}
\centering
\resizebox{\textwidth}{!}{
\begin{tabular}{l|cccccc}
\toprule
Model & Training Time (s) & Peak Mem (GB) & Params (MB) & FLOPs (GF) & CO$_2$ Emission (kg) \\
\midrule
Art         & 1559.0 & 0.81 & 363.7 & 70.7 & 0.015 \\
Clipart     & 1559.0 & 0.81 & 363.7 & 70.7 & 0.023 \\
Product     & 1569.5 & 0.81 & 363.7 & 70.7 & 0.023 \\
Real World & 1554.8 & 0.81 & 363.7 & 70.7 & 0.023 \\
\midrule
\textbf{Average (Teacher)} & \textbf{1560.6} & \textbf{0.81} & \textbf{363.7} & \textbf{70.7} & \textbf{0.021} \\
\midrule
\textbf{Student (Target)}  & \textbf{2094.2} & \textbf{1.25} & \textbf{363.7} & \textbf{70.7} & \textbf{0.051} \\
\bottomrule
\end{tabular}
}
\end{table}

\subsection{Limitations and Failure Modes}
\begin{figure}[htbp]
\centering
\includegraphics[width=0.7\textwidth]{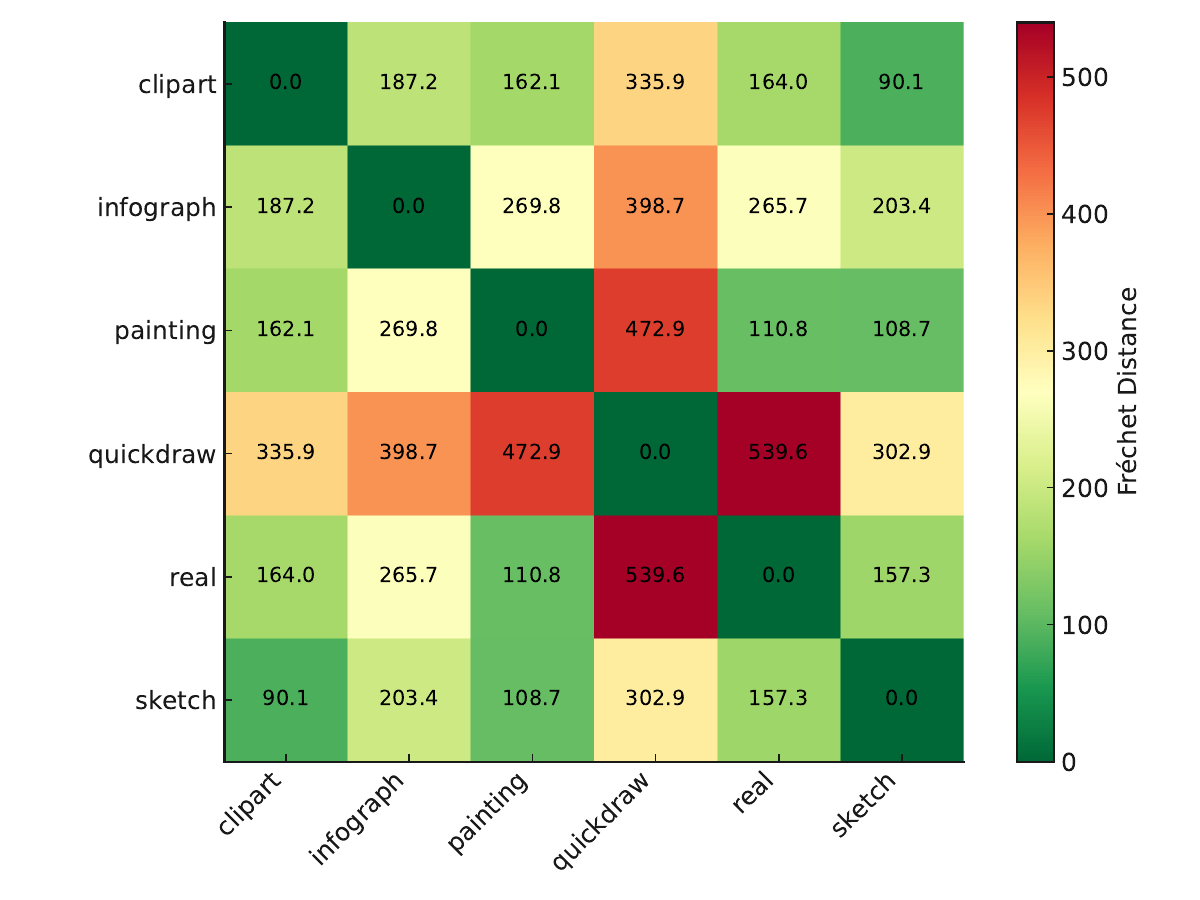}
\caption{Fréchet distances between DomainNet domains.}
\label{fig:frechet}
\end{figure}

Although SKD-ETR performs competitively across most domains, certain cases reveal some limitations of its design. The Quickdraw domain in DomainNet poses a particular challenge due to its extreme stylization and substantial distribution gap from the sources. In Figure \ref{fig:frechet}, we provide a quantitative measure of the domain shift by computing the pairwise Fréchet Distance between all domains in DomainNet using embeddings extracted from the frozen DINOv2 backbone. As illustrated, the Quickdraw domain exhibits the largest distances relative to all others, forming a clear outlier in the overall domain structure.

SKD-ETR employs random initialization to avoid bias toward any specific source; thus, the student begins without domain-aligned priors. While this unbiased start generally enhances robustness and fairness, it can hinder adaptation when the target distribution differs drastically from all sources. As evident from the Quickdraw results, performance declines relative to aggregation-based approaches such as SEA+MSPL \cite{abedi_one-shot_2025}, which benefit from source-informed initialization. Nevertheless, SKD-ETR attains the best average performance overall, demonstrating strong generalization ability. Future work will investigate adaptive initialization or progressive distillation strategies to better handle extreme domain shifts.

\section{Conclusion and Future Work}
\label{sec:conclusion}

In this paper, we introduced SKD-ETR, a novel one-shot FUDA framework that effectively transfers knowledge from multiple decentralized source models to an unlabeled target domain without requiring access to source data. Our method consists of two key modules: SKD, which regularizes training by mitigating the effects of noisy pseudo labels using label smoothing; and ETR, which updates the source models after each iteration of student updates using the student’s parameters. This process enables the teachers to generate increasingly better and more robust pseudo-labels after each iteration, without requiring additional communication. We also provide a theoretical justification for the EMA update and show that our framework avoids complex aggregation mechanisms.
Extensive experiments on OfficeHome, Office-Caltech, and DomainNet benchmarks demonstrate that SKD-ETR achieves superior or comparable performance to state-of-the-art methods while requiring significantly fewer trainable parameters and communication rounds. These results validate the effectiveness and the communication and computation efficiency of our approach in practical federated settings.
For future work, we plan to extend our framework to more complex scenarios such as heterogeneous model architectures and semi-supervised target domains. Moreover, exploring adaptive smoothing strategies and integrating model compression techniques may further enhance scalability and deployment in resource-constrained environments.

\section*{Acknowledgment}
This research is partially funded by the NSERC CREATE TrustCAV program and the NSERC Discovery Grant program, and in part by the NSERC Canada Research Chair (CRC) Program.

\bibliographystyle{elsarticle-num} 
\bibliography{references}

\end{document}